\documentclass[review]{elsarticle}

\usepackage{lineno,hyperref}
\usepackage[top=1in,bottom=1in,left=0.75in,right=0.75in]{geometry}
\usepackage[usenames, dvipsnames]{color}
\usepackage{amsmath}
\usepackage{graphicx}
\usepackage{subfigure}
\usepackage{amssymb}
\numberwithin{equation}{section}
\usepackage{subfigure}
\usepackage{threeparttablex}
\usepackage{csquotes}
\usepackage{bm}
\newcommand{\vect}[1]{\boldsymbol{#1}} 
\usepackage{enumitem}
\usepackage{blindtext}
\usepackage{scrextend}
\addtokomafont{labelinglabel}{\sffamily}
\usepackage{epstopdf}
\usepackage{caption}
\usepackage{array}
\usepackage{color,colortbl}
 
\modulolinenumbers[5]

\journal{Journal of \LaTeX\ Templates}

\begin{document}

\begin{frontmatter}

\title{The Chan-Vese Model with Elastica and Landmark Constraints for Image Segmentation}

\author[mymainaddress]{Jintao Song}
\ead[url]{2017021234@qdu.edu.cn}
\fntext[myfootnote]{Since 1880.}

\author[mysecondaryaddress]{Huizhu Pan}
\ead[url]{huizhu.pan@postgrad.curtin.edu.au}

\author[mysecondaryaddress]{Wuanquan Liu}
\ead[url]{w.liu@curtin.edu.au}

\author[mythirdaddress]{Zisen Xu}
\ead[url]{zisen\_xu@126.com}

\author[mymainaddress]{Zhenkuan Pan}
\cortext[mycorrespondingauthor]{Zhenkuan Pan}
\ead{zkpan@126.com}

\address[mymainaddress]{College of Computer Science and Technology, Qingdao University, \\
	Qingdao, 266071, China}
\address[mysecondaryaddress]{School of Electrical Engineering, Mathematical Science and Computing,\\
	Curtin University, Perth, WA 6102, Australia}
\address[mythirdaddress]{The Affiliated Hospital of Qingdao University,\\
	Qingdao, 266003, China}

\begin{abstract}
\noindent In order to completely separate objects with large sections of occluded boundaries in an image, we devise a new variational level set model for image segmentation combining the Chan-Vese model with elastica and landmark constraints. For computational efficiency, we design its Augmented Lagrangian Method (ALM) or Alternating Direction Method of Multiplier (ADMM) method by introducing some auxiliary variables, Lagrange multipliers, and penalty parameters. In each loop of alternating iterative optimization, the sub-problems of minimization can be easily solved via the Gauss-Seidel iterative method and generalized soft thresholding formulas with projection, respectively. Numerical experiments show that the proposed model can not only recover larger broken boundaries but can also improve segmentation efficiency, as well as decrease the dependence of segmentation on parameter tuning and initialization.\\
\noindent
\end{abstract}


\end{frontmatter}

\linenumbers

\section{Introduction}
Variational level set methods\cite{zhao1996variational} have been widely applied to image segmentation problems based on image features such as edge, region, texture and motion, etc.\cite{osher2003geometric,mumford1989optimal,chan2005image,mitiche2010variational}. For images containing occluded objects, variational models using shape priors can inpaint missing boundaries based on pre-defined shapes\cite{chen2002using,cremers2008nonlinear,chen2012image,thiruvenkadam2008segmentation}. However, obtaining shape priors is often not easy. In certain scenarios, using landmarks is a good alternative to using shape priors in the segmentation of occluded objects.
Motivated by image registration with landmarks\cite{modersitzki2004numerical,goshtasby2012image,lam2014landmark}, Pan et al.\cite{pan2019novel} proposed a Chan-Vese model\cite{chan2001active} with landmark constraints (CVL) under the variational level set framework. The model not only enforces the segmentation contour to pass through some pre-selected feature points but also improves computational efficiency and weakens the dependence of the segmentation result on initialization.  However, since the Chan-Vese model uses the total variation (TV) \cite{rudin1992nonlinear} of the Heaviside function of the level set function to approximate the length of contours, sometimes the details are not handled well enough. So the CVL model performs better for recovering smaller boundaries. On the other hand, the elastica regularizers proposed in the early 1990s in depth segmentation\cite{mumford1994elastica} have been successively applied to image inpainting with larger broken images\cite{shen2003euler}, image restoration with smooth components\cite{zhu2012image,zhu2013augmented}, and image segmentation with larger damaged areas or occlusions\cite{zhu2013image,zhu2006segmentation,zhang2016new,tan2018image}. In \cite{zhu2013image}, Zhu et al. propose a modified Chan-Vese model with elastica (CVE), combining the classic Chan-Vese model (CV) and the elastica regularizer to inpaint, or interpolate, segmentation curves. Due to the ill-posed nature of this model, the segmentation results rely heavily on the involved penalty parameters, which makes it hard to for curves to pass through desired points in the occluded regions. In this paper, we propose a CVE model with landmark constraints (CVEL) that combines the CVL and CVE to more accurately and robustly complete missing curves. Different from the CVE proposed in \cite{zhu2013image} which uses piecewise constant level set functions or binary label functions, we use the Lipschitz smooth level set function defined as a signed distance function to describe curve evolution. In order to solve the proposed model with the signed distance property and landmark constraints, we devise its Augmented Lagrangian Method (ALM), i.e., Alternating Direction Method of Multipliers  (ADMM)\cite{wu2010augmented,tai2011fast,goldstein2014fast,duan2017introducing}  solution by dividing the original problem into several simple sub-problems and optimizing them alternatively. The sub-problems can be solved respectively by the Gauss-Seidel iterative method and generalized soft thresholding formula with projection \cite{duan2014some}.

The paper is organized as follows. In section 2, we present the classical Chan-Vese model, the Chan-Vese model with elastica, and the Chan-Vese model with landmark constraints for comparisons consequently. In section 3, we give the CVE model with landmark constraints under the variational level set framework and design its ADMM method. Section 4 covers the solutions to all sub-problems derived in section 3. Numerical examples are presented in section 5 to show the performance of the proposed model and algorithm. Finally, concluding remarks are drawn in section 6.    

\section{The Previous Works}
\label{TWSO}

\subsection{The Chan-Vese model for image segmentation}
The task of two-phase segmentation of a gray value image $f\left(x\right) : \Omega \rightarrow R$ is to divide $\Omega$ into two regions $\Omega_1,\Omega_2$,  such that $\Omega = \Omega_1 \bigcap \Omega_2$ and $\Omega_1 \bigcap \Omega_2 \neq \varnothing$. The classical Chan-Vese model\cite{chan2001active} is a reduced piecewise constant Mumford-Shah model\cite{mumford1989optimal} under the variational level set framework. The original image is denoted as $f\left(x\right) = c_1 \chi_1\left(\phi\left(x\right)\right)+c_2 \chi_2\left(\phi\left(x\right)\right)$, where $c_1$ and $c_2$ are the average image intensities in $\Omega_1 ,\Omega_2$, and $\chi_1\left(\phi\left(x\right)\right)=H\left(\phi\left(x\right)\right)\in\left[0,1\right]$ and $\chi_2\left(\phi\left(x\right)\right)=1-H\left(\phi\left(x\right)\right)\in\left[0,1\right]$ are characteristic functions of $\Omega_1,\Omega_2$ respectively. $\phi\left(x\right)$ is a level set function defined as a signed distance function form point $x$ to curve $\Gamma$, i. e. 
\begin{equation}
\phi\left(x\right)= \left\{ \begin{array}{rcl}
&d\left(x,\Gamma\right),&\text{if}\;x \in \Omega_1\\
&0,&\text{if}\;x \in \Gamma\\
&-d\left(x,\Gamma\right),&\text{if}\;x \in \Omega_2\\
\end{array}\right.,
\end{equation}
with the property
\begin{equation}
|\nabla \phi\left(x\right)|=1.
\end{equation}
(2.2) is the Eikonal equation, i. e., a kind of Hamilton-Jacobi equation. $H\left(\phi\left(x\right)\right)$ is the Heaviside function of $\phi\left(x\right)$, stated as 
\begin{equation}
H\left(\phi\left(x\right)\right)= \left\{ \begin{array}{rcl}
&1,& \text{if}\; \phi\left(x\right) \geq 0 \\
&0,& \text{otherwise}
\end{array}\right.,
\end{equation}
its partial derivative with $\phi\left(x\right)$ is the Dirac function  
\begin{equation}
\delta\left(\phi\right)=\frac{\partial H\left(\phi\right)}{\partial\phi},
\end{equation}
which is a generalized function. Usually, $H\left(\phi\right)$, $\delta\left(\phi\right)$ are replaced by their mollified versions by introducing a small positive constant parameter $\varepsilon$ , for instance \cite{chan2001active}\\    
\begin{equation}
H_\varepsilon\left(\phi\right)=\frac{1}{2}\left(1+\frac{2}{\pi}arctan\left(\frac{\phi}{\varepsilon}\right)\right),
\end{equation}  
\begin{equation}
\delta_\varepsilon\left(\phi\right)=\frac{\partial H_\varepsilon\left(\phi\right)}{\partial\phi}=\frac{1}{\pi}\frac{\varepsilon^2}{\varepsilon^2+\phi^2}.
\end{equation}  
The well-known Chan-Vese model\cite{chan2001active} for two-phase image segmentation is an energy minimization problem on $c_1$, $c_2$ and $\phi$, such that
\begin{equation}
\text{min}E(c_1,\;c_2,\;\phi) =
\int_\Omega \left(f-c_1\right)^2 H_\varepsilon \left(\phi\right)dx+\int_\Omega \left(f-c_2\right)^2 \left(1-H_\varepsilon \left(\phi\right)\right)dx
+\gamma \int_\Omega|\nabla H_\varepsilon \left(\phi\right)|dx,
\;\; s.t. \; |\nabla \phi|=1.
\end{equation}
Where $\gamma$ is a penalty parameter for the length term of the curve. 

\noindent$c_1$ and $c_2$ are estimated as
\begin{equation}
{c_1}=\frac{{\int_\Omega f\left(x\right)H_\varepsilon \left(\phi\left(x\right)\right)dx}}{ \int_\Omega {H_\varepsilon \left(\phi\left(x\right)\right)dx}},
\end{equation}
\begin{equation}
{c_2}=\frac{ {\int_\Omega f\left(x\right)\left(1-H_\varepsilon \left(\phi\left(x\right)\right)\right)dx}}{ \int_\Omega {\left(1-H_\varepsilon \left(\phi\left(x\right)\right)\right)dx}}.
\end{equation}
By introducing $Q(c_1,\; c_2)=\alpha_1 (c_1-f)^2-\alpha_2 (c_2-f)^2$, (2.7) can be rewritten as 
\begin{equation}
\text{min}E(c_1,\;c_2,\;\phi) =\int_\Omega Q\left(c_1,\;c_2\right)H_\varepsilon \left(\phi\right)dx +\gamma \int_\Omega|\nabla H_\varepsilon \left(\phi\right)|dx,\;\; s.t. \; |\nabla \phi|=1.
\end{equation}
Therefore, the evolution equation of $\phi\left(x\right)$ can be derived via variational methods and gradient descent as
\begin{equation}
\left\{ \begin{split}
&\frac{\partial \phi\left(x,\;t\right)}{\partial t}=\left(\nabla \cdot \left(\frac{\nabla \phi\left(x,\;t\right)}{|\nabla \phi\left(x,\;t\right)|}\right)-Q\left(c_1,c_2\right)\right)\delta_\varepsilon\left(\phi\left(x,\;t\right)\right) 
&t>0,\;x\in \Omega\\
&\frac{\partial \phi\left(x,\;t\right)}{\partial N} =0 
&t>0,\;x\in \partial\Omega\\
&\phi\left(x,\;0\right)=\phi^0\left(x\right) 
&t=0,\;x\in \Omega
\end{split}\right.,
\end{equation}

\subsection{The Chan-Vese model with landmark constraints\cite{pan2019novel}}    
Let $x_L=\{x_1,x_2...x_l\}$ be the given landmark points, represented through a mask function
\begin{equation}
\eta \left(x\right)= \left\{ \begin{array}{rcl}
&1,\quad & \text{if}\;x \in x_L \\
&0,\quad & \text{otherwise}
\end{array}\right..
\end{equation}
Since the zero level set describes the boundary curve and the landmarks are positioned on the boundary, the landmark constraint is
\begin{equation}
\phi \left(x\right)=0,\;\text{if} \; \eta\left(x\right)=1.
\end{equation}
Thus, the Chan-Vese model (2.9) can be transformed into the following constrained optimization problem
\begin{equation}
\text{min}E(c_1,\;c_2,\;\phi) =
\int_\Omega Q\left(c_1,\;c_2\right)H_\varepsilon (\phi) dx
+\gamma \int_\Omega|\nabla H_\varepsilon \left(\phi\right)|dx
+\frac{\mu}{2}\int_\Omega \eta\phi^2,
\;\; s.t. \; |\nabla \phi|=1,
\end{equation}
where $\mu >0$ is the penalty parameter.

\subsection{The Chan-Vese model with elastica}
In order to recover curves which are not determined by image features, for instance the boundary of an occluded object, \cite{zhu2013image,tai2017simple} proposed the CVE model by combining Chan-Vese model and the elastica term
\begin{equation}
\text{min}E(c_1,\;c_2,\;\phi) =
\int_\Omega Q\left(c_1,\;c_2\right)H_\varepsilon (\phi) dx
+\gamma \int_\Omega\left(a+b\left(\nabla \cdot\left(\frac{\nabla \phi}{|\nabla \phi|}\right)\right)^2\right)|\nabla H_\varepsilon \left(\phi\right)|dx,
\;\; s.t. \; |\nabla \phi|=1,
\end{equation}
where $(\nabla \cdot\left(\frac{\nabla \phi}{|\nabla \phi|}\right))^2$ is the elastica, i. e. the square of the curvature. The contours obtained by this method tend to be curved rather than straight. Later, Zhu et al. \cite{zhu2013image} considered the relation
\begin{equation}
\frac{\nabla H_\varepsilon\left(\phi\right)}{|\nabla H_\varepsilon\left(\phi\right)|}=\frac{\nabla \phi \delta\left(\phi\right)}{|\nabla \phi| \delta\left(\phi\right)}=\frac{\nabla \phi}{|\nabla \phi|}.
\end{equation}
and studied the following convex optimization problem instead of (2.15) 
\begin{equation}
\text{min}E(c_1,\;c_2,\;\phi) =
\int_\Omega Q\left(c_1,\;c_2\right)\phi dx
+\gamma \int_\Omega\left(a+b\left(\nabla \cdot\left(\frac{\nabla \phi}{|\nabla \phi|}\right)\right)^2\right)|\nabla \phi|dx,
\;\; s.t. \; |\nabla \phi|=1.
\end{equation}
However, the investigations of this paper are based solely on (2.15). Because the reason for the simplification is to facilitate the calculation, and will not affect the experimental results.
\section{The CVE model with landmark constraints and its ADMM algorithm}
Combining (2.14) and (2.15), we propose the Chan-Vese model with elastica and landmark as
\begin{equation}
\begin{split}
\text{min}E(c_1,\;c_2,\;\phi)
&=\int_\Omega Q\left(c_1,\;c_2\right)H_\varepsilon (\phi) dx
+\frac{\mu}{2}\int_\Omega \eta\phi^2dx\\
&+\gamma \int_\Omega\left(a+b\left(\nabla \cdot\left(\frac{\nabla \phi}{|\nabla \phi|}\right)\right)^2\right)|\nabla H_\varepsilon \left(\phi\right)|dx, \\
&s.t. \; |\nabla \phi|=1.
\end{split}
\end{equation}
By adding landmark points, you can force the contour to pass through some feature points to get good results. This is the design idea of The CVE model with landmark (CVEL). In order to simplify the implementation of (3.1), we introduce auxiliary variables $\bm{p},\bm{m},\bm{n},q$, The main reason for adding so many intermediate quantities is to avoid the curvature term appearing in the calculation and to simplify the calculation. Those are such that 
\begin{equation}
\bm{p}=\nabla \phi,
\end{equation}
\begin{equation}
\bm{m}=\frac{\bm{p}}{|\bm{p}|},
\end{equation}
\begin{equation}
q=\nabla \cdot \bm{n}.
\end{equation}
Considering that $|\bm{m}| \leq 1$, (3.3) can be substituted by a more relaxed set of constraints, $\;$$|\bm{p}|-\bm{p}\cdot \bm{m} \leq 1$ and $|\bm{m}| \leq 1$ \cite{mumford1989optimal}. Since $\bm{p}= \nabla \phi$, the constraint $ |\nabla \phi|=1$ can be rewritten as $|\bm{p}|=1$. Additionally, we introduce a new variable $\bm{n}=\bm{m}$ \cite{mumford1989optimal} for splitting. Thus, the constraints (3.2)-(3.4) can be summarized as
\begin{equation}
\bm{p}=\nabla \phi,
\end{equation}
\begin{equation}
|\bm{p}|-\bm{p}\bm{m} = 0,|\bm{p}|=1,
\end{equation}
\begin{equation}
\bm{n}=\bm{m},|\bm{m}|\leq 1,
\end{equation}
\begin{equation}
q=\nabla \cdot \bm{n}.
\end{equation}
Next, to design the ADMM algorithm for the problem, we introduce the Lagrange multipliers $\lambda_1,\bm{\lambda_2},\bm{\lambda_3},\lambda_4$ and penalty parameters $\gamma_1,\gamma_2,\gamma_3,\gamma_4$ and rewrite the energy function in (3.1) as the following Augmented Lagrangian Function
\begin{equation}
\begin{split}
E\left(c_1,c_2,u,\phi,\bm{v},\bm{p},\bm{n},\bm{m}\right)
&=\int_\Omega Q\left(c_1,\;c_2\right)H\left(\phi\right) dx \\
&+\gamma \int_\Omega\left(a+bq^2\right)|\bm{p}|\delta_\varepsilon\left(\phi\right)dx+\frac{\mu}{2}\int_\Omega \eta\phi^2dx\\
&+\int_\Omega \lambda_1\left(|\bm{p}|-\bm{p}\cdot\bm{m}\right)dx
+\gamma_1\int_\Omega\left(|\bm{p}|-\bm{p}\cdot\bm{m}\right)dx\\
&+\int_\Omega \bm{\lambda_2}\left(\bm{p}-\nabla \phi\right)dx
+\frac{\gamma_2}{2}\int_\Omega|\bm{p}-\nabla \phi|^2dx\\
&+\int_\Omega \bm{\lambda_3}\left(\bm{n}-\bm{m}\right)dx
+\frac{\gamma_3}{2}\int_\Omega\left(\bm{n}-\bm{m}\right)^2dx + \delta_{\cal {R}}(\vect{m})\\
&+\int_\Omega \lambda_4\left(q-\nabla\cdot\bm{n}\right)dx
+\gamma_4\int_\Omega\left(q-\nabla\cdot\bm{n}\right)^2dx\\
\end{split},
\end{equation}
where $ \left|\bm{p}\right|=1,{\cal {R}}= \left\{ {m \in {L^2}\left( \Omega  \right):|m| \le 1\;\text{a.e.}\;\text{in}\;\Omega } \right\}$ and $\delta_{\cal {R}}(\vect{m})$ is the characteristic function on the convex set ${\cal {R}}$, given by
\[\delta_{\cal {R}}(\vect{m}) = \left\{ \begin{array}{ll}
0 & \text{if}\; \vect{m} \in {\cal {R}}\\
+ \infty  & \text{otherwise}
\end{array} \right..\]
Under the framework of ADMM, the Lagrangian multipliers are updated for iteration $k=0,1,2...K$ as
\begin{equation}
\left\{ \begin{split}
&\lambda_1^{k+1}=\lambda_1^{k}+\gamma_1\left(|\bm{p}^{k+1}|-\bm{p}^{k+1}\cdot\bm{m}^{k+1}\right)\\
&\bm{\lambda_2}^{k+1}=\bm{\lambda_2}^{k}+\gamma_2\left(\bm{p}^{k+1}-\nabla \phi^{k+1}\right)\\
&\bm{\lambda_3}^{k+1}=\bm{\lambda_3}^{k}+\gamma_3\left(|\bm{n}^{k+1}|-\bm{m}^{k+1}\right)\\
&\lambda_4^{k+1}=\lambda_4^{k}+\gamma_4\left(q^{k+1}-\nabla\cdot\bm{n}^{k+1}\right)
\end{split}\right.,
\end{equation}
and the original minimization problem is split into the following sub-problems
\begin{equation}
c_1^{k+1}=\text{arg} \min\limits_{c_1}E\left(c_1, c_2^{k},\;\phi^{k},\;\bm{p}^{k},\;\bm{n}^{k},\;\bm{m}^{k},\;q^{k}\right),
\end{equation}
\begin{equation}
c_2^{k+1}=\text{arg} \min\limits_{c_2}E\left(c_1^{k+1},\;c_2,\phi^{k},\;\bm{p}^{k},\;\bm{n}^{k},\;\bm{m}^{k},q^{k}\right),
\end{equation}
\begin{equation}
\phi^{k+1}=\text{arg} \min\limits_{\phi}E\left(c_1^{k+1},\;c_2^{k+1},\;\phi,\;\bm{p}^{k},\;\bm{n}^{k},\;\bm{m}^{k},\;q^{k}\right),
\end{equation}
\begin{equation}
\bm{p}^{k+1}=\text{arg} \min\limits_{\bm{p}}E\left(c_1^{k+1},\;c_2^{k+1},\;\phi^{k+1},\;\bm{p},\;\bm{n}^{k},\;\bm{m}^{k},\;q^{k}\right),
\end{equation}
\begin{equation}
\bm{n}^{k+1}=\text{arg} \min\limits_{\bm{n}}E\left(c_1^{k+1},\;c_2^{k+1},\;\phi^{k+1},\;\bm{p}^{k+1},\;\bm{n},\;\bm{m}^{k},\;q^{k}\right),
\end{equation}
\begin{equation}
\bm{m}^{k+1}=\text{arg} \min\limits_{\bm{m}}E\left(c_1^{k+1},\;c_2^{k+1},\;\phi^{k+1},\;\bm{p}^{k+1},\;\bm{n}^{k+1},\;\bm{m},\;q^{k}\right),
\end{equation}
\begin{equation}
q^{k+1}=\text{arg} \min\limits_{q}E\left(c_1^{k+1},\;c_2^{k+1},\;\phi^{k+1},\;\bm{p}^{k+1},\;\bm{n}^{k+1},\;\bm{m}^{k+1},\;q\right).
\end{equation}
The solutions to the sub-problems are presented below. 

Using standard variational methods, we solve (3.11) and (3.12) respectively and get
\begin{equation}
c_1^{k+1}=\frac{\int_\Omega f\left(x\right) H\left(\phi^k\left(x\right)\right) dx}{\int_\Omega H\left(\phi^k\left(x\right)\right) dx},
\end{equation}
\begin{equation}
c_2^{k+1}=\frac{\int_\Omega f\left(x\right)\left(1-H\left(\phi^k\left(x\right)\right)\right) dx}{\int_\Omega \left(1-H\left(\phi^k\left(x\right)\right)\right) dx}.
\end{equation}

For the sub-problem (3.13), the Euler-Lagrange equations on $\phi$ are 
\begin{equation}
\left\{\begin{split}
&F^{k+1}+ \mu \eta \phi^{k+1}-\gamma _2 \Delta \phi^{k+1}=0&{x \in \Omega }\\
&{\left( { - \bm \lambda _2^k + {\gamma _2}\left( {\nabla \phi^{k+1}  - {{\bm p}^k}} \right)} \right) \cdot \bm N = 0}&{x \in \partial \Omega }
\end{split}\right..
\end{equation}
where, $F^{k+1}={{Q^{k + 1}}{\delta _\varepsilon }\left( \phi^{k}  \right) + \left( {a + b{q^2}} \right)\left| {\bm{p}^k} \right|\nabla {\delta _\varepsilon }\left( \phi^{k}  \right)+\nabla  \cdot \bm \lambda _2^k + {\gamma _2}\nabla  \cdot  {{\bm p}^k}}$.

To solve (3.14), we can derive $\bm{p}^{k+1}$ via a generalized soft thresholding formula and projection formula as below
\begin{equation}
\left\{\begin{split}
&{{\bm A}^{k + 1}} = \nabla {\phi ^{k + 1}} + \frac{{\left( {\lambda_1^k + {\gamma _1}} \right){{\bm m}^k} - \bm \lambda _2^k}}{{{\gamma _2}}}\\
&{B^{k + 1}} = \left( {a + b{{\left( {{q^k}} \right)}^2}} \right)\nabla {\delta _\varepsilon }\left( {{\phi ^k}} \right)\\
&{{\bm{\tilde p}}^{k + 1}} = \text{max}\left( {\left| {{{\bm A}^{k + 1}}} \right| - \frac{{\lambda _1^k + {\gamma _1} + {B^{k + 1}}}}{{{\gamma _2}}},0} \right)\frac{{{{\bm A}^{k + 1}}}}{{\left| {{{\bm A}^{k + 1}}} \right| + {{10}^{ - 6}}}}\\
&{{\bm p}^{k + 1}} = \frac{{{{\bm{\tilde p}}^{k + 1}}}}{{\left| {{{\bm {\tilde p}}^{k + 1}}} \right|}},\;\frac{{\bm 0}}{{\left| {\bm 0} \right|}} = \bm 0
\end{split}\right..
\end{equation}

For (3.15), the Euler-Lagrange equations on $\bm{n}$ is
\begin{equation}
\bm \lambda _3^k + {\gamma _3}\left( {\bm{n}^{k+1} - {{\bm m}^k}} \right) + {\gamma _4}\nabla \left( {q^k - \nabla  \cdot {{\bm n }^k}} \right) + \nabla {\lambda _4^k} = 0.
\end{equation}

$\bm m$ in (3.16) can be obtained as an exact solution. Considering the constraint in (3.7), the projection formula should be augmented as follows
\begin{equation}
\left\{ \begin{split}
&\bm{\widetilde{m}}^{k+1}=\bm{n}^{k+1}+\frac{{\left( {\lambda _1^k + {\gamma _1}} \right){{\bm p}^{k + 1}} + \bm \lambda _3^k}}{{{\gamma _3}}}\\
&\bm{m}^{k+1}=\frac{\bm{\widetilde{m}}^{k+1}}{\text{max}\left(1,|\bm{\widetilde{m}}^{k+1}|\right)}
\end{split}\right..
\end{equation}

Lastly, $q$ in (3.17) also has an analytical solution
\begin{equation}
{\gamma _4}\left( {q^{k+1} - \nabla  \cdot {{\bm n }^{k + 1}}} \right) + {\lambda _4}^k + 2bq^{k+1}\left| {\bm{p}^{k+1} } \right|{\delta _\varepsilon }\left( \phi^{k+1}  \right) = 0.
\end{equation}

In this part we introduce the solution of each variable in the CVEL model and the iterative process of the algorithm. In the next section we will combine the experiments to introduce the role of the model in detail.
\section{Implementations of the relevant sub-problems of minimization}
To compute (3.18)-(3.24) and (3.10) numerically, we need to design discrete algorithms for the sub-problems. For the sake of simplicity, we discretize the image domain pixel by pixel with the rows and column numbers as indices. Then, the gradients can be represented approximately by forward, backward and central finite differences
\begin{equation}
{\nabla ^ + }{\phi _{i,j}} = \left[ \begin{array}{l}
\partial _{{x_1}}^ + {\phi _{i,j}}\\
\partial _{{x_2}}^ + {\phi _{i,j}}
\end{array} \right],\;{\nabla ^ - }{\phi _{i,j}} = \left[ \begin{array}{l}
\partial _{{x_1}}^ - {\phi _{i,j}}\\
\partial _{{x_2}}^ - {\phi _{i,j}}
\end{array} \right],\;{\nabla ^o}{\phi _{i,j}} = \left[ \begin{array}{l}
\partial _{{x_1}}^o{\phi _{i,j}}\\
\partial _{{x_2}}^o{\phi _{i,j}}
\end{array} \right],
\end{equation}
where, 
\begin{equation}
\left\{ \begin{split}
&\partial _{{x_1}}^ + {\phi _{i,j}} = {\phi _{i + 1,j}} - {\phi _{i,j}}\\
&\partial _{{x_1}}^ - {\phi _{i,j}} = {\phi _{i,j}} - {\phi _{i - 1,j}}
\end{split} \right.,\;\left\{ \begin{split}
&\partial _{{x_2}}^ + {\phi _{i,j}} = {\phi _{i,j + 1}} - {\phi _{i,j}}\\
&\partial _{{x_2}}^ - {\phi _{i,j}} = {\phi _{i,j}} - {\phi _{i,j - 1}}
\end{split} \right.,\;\left\{ \begin{split}
&\partial _{{x_1}}^o{\phi _{i,j}} = \frac{1}{2}\left( {{\phi _{i + 1,j}} - {\phi _{i - 1,j}}} \right)\\
&\partial _{{x_2}}^o{\phi _{i,j}} = \frac{1}{2}\left( {{\phi _{i,j + 1}} - {\phi _{i,j - 1}}} \right)
\end{split} \right..
\end{equation}
The discretized Laplacian of $\phi$ can be stated as
\begin{equation}
\Delta {\phi _{i,j}} = {\nabla ^ - } \cdot \left( {{\nabla ^ + }{\phi _{i,j}}} \right) = {\phi _{i - 1,j}} + {\phi _{i,j - 1}} + {\phi _{i + 1,j}} + {\phi _{i,j + 1}} - 4{\phi _{i,j}}.
\end{equation}
The other variables can be expressed in similar ways.

(3.18) and (3.19) can be calculated directly as

\begin{equation}
c_1^{k + 1} = \frac{{\sum\limits_{i = 1}^M {\sum\limits_{j = 1}^N {{f_{i,j}}} } H\left(\phi_{i,j}^k\right)}}{{\sum\limits_{i = 1}^M {\sum\limits_{j = 1}^N {H\left(\phi_{i,j}^k\right)} } }},
\end{equation}
\begin{equation}
c_2^{k + 1} = \frac{{\sum\limits_{i = 1}^M {\sum\limits_{j = 1}^N {{f_{i,j}}\left( {1 - H\left(\phi_{i,j}^k\right)} \right)} } }}{{\sum\limits_{i = 1}^M {\sum\limits_{j = 1}^N {\left( {1 - H\left(\phi_{i,j}^k\right)} \right)} } }},
\end{equation}
where M and N are the numbers of rows and columns of the image $f$.

Next, to discretize the formula of $\phi$ obtained in (3.20), we introduce the following intermediate variables
\begin{equation}
\left\{ {\begin{split}
	&{{F^{k+1}} = {Q^{k + 1}}{\delta _\varepsilon }\left( \phi^k  \right) + \left( {a + b{\left(q^k\right)^2}} \right)\left| {\bm p}^k \right|\nabla {\delta _\varepsilon }\left( \phi^k  \right) + {\gamma _2}\nabla  \cdot \bm p^k + \nabla  \cdot {{\bm \lambda }_2^k}}&{x \in \Omega }\\
	&{{{\bm G}^{k+1}} = {{\bm p }^k} + \frac{{\bm \lambda _2^k}}{{{\gamma _2}}}}&{x \in \partial \Omega }
	\end{split}} \right.,
\end{equation}
and write the original Eular-Lagrange equations in the more concise form below
\begin{equation}
\left\{ {\begin{split}
	&{{F^{k+1}} + \mu \eta \phi^{k+1}  - {\gamma _2}\Delta \phi^{k+1}  = 0}&{x \in \Omega }\\
	&{\nabla \phi^{k+1}  \cdot \bm N = {{\bm G}^{k+1}} \cdot \bm N}&{x \in \partial \Omega }
	\end{split}} \right..
\end{equation}

Based on (4.3) and (4.7), we can easily design the Gauss-Seidel iterative scheme of $\phi$ as
\begin{equation}
\left( {\mu \eta  + 4{\gamma _2}} \right)\phi _{i,j}^{k + 1,l + 1} = {\gamma _2}\left( {\phi _{i - 1,j}^{k + 1,l + 1} + \phi _{i,j - 1}^{k + 1,l + 1} + \phi _{i + 1,j}^{k + 1,l} + \phi _{i,j + 1}^{k + 1,l}} \right) - F_{i,j}^{k+1}.
\end{equation}

Alternatively, $\phi$ can be solved by Fast Fourier transform (FFT)\cite{duan2017introducing}.

The discretized solution of $\bm{p}$ as obtained from (3.21) is
\begin{equation}
\left\{ \begin{split}
&\bm A_{i,j}^{k + 1} = {\nabla ^ + }\phi _{i,j}^{k + 1} + \frac{{\left( {\lambda _1^k + {\gamma _1}} \right)\bm m_{i,j}^k - \bm \lambda _{2i,j}^k}}{{{\gamma _2}}}\\
&{B^{k + 1}} = \left( {a + b{{\left( {q_{i,j}^k} \right)}^2}} \right){\nabla ^ + }{\delta _\varepsilon }\left( {\phi _{i,j}^{k + 1}} \right)\\
&\bm {\tilde p_{i,j}}^{k + 1} = \text{max}\left( {\left| {\bm A_{i,j}^{k + 1}} \right| - \frac{{\lambda _{1i,j}^k + {\gamma _1} + B_{i,j}^{k + 1}}}{{{\gamma _2}}},0} \right)\frac{{\bm A_{i,j}^{k + 1}}}{{\left| {\bm A_{i,j}^{k + 1}} \right| + {{10}^{ - 6}}}}\\
&\bm p_{i,j}^{k + 1} = \frac{{\bm{ \tilde p_{i,j}}^{k + 1}}}{{\left| {\bm {\tilde p_{i,j}}^{k + 1}} \right|}},\;\frac{{0}}{{\left| {\bm 0} \right|}} = \bm0
\end{split} \right..
\end{equation}

Since the form of $\bm{n}$ in (3.22) is similar to that of $\phi$, the solution of $\bm{n}$ can also be written similarly. Again, to simplify the equation, we introduce 
\begin{equation}
\left\{ {\begin{split}
	&{F_{i,j}^{k + 1} = \bm \lambda _{3i,j}^k + {\gamma _4}\nabla q_{i,j}^k + \nabla \lambda _{4i,j}^k - {\gamma _3}\bm m_{i,j}^k}&{x \in \Omega }\\
	&{{{\bm G}^k} = q_{i,j}^k + \frac{{\lambda _{4i,j}^k}}{{{\gamma _4}}}}&{x \in \partial \Omega }
	\end{split}} \right.,
\end{equation}
and (3.22) becomes
\begin{equation}
\left\{ {\begin{split}
	&{{F^k} + {\gamma _3}\bm{n}^{k+1}  - {\gamma _4}\nabla\cdot\left(\nabla \bm{n}^{k+1} \right)   = 0}&{x \in \Omega }\\
	&{\nabla \bm{n}^{k+1}   \cdot \bm N = {{\bm G}^k} \cdot \bm N}&{x \in \partial \Omega }
	\end{split}} \right..
\end{equation}
Introducing the discretized form of $\bm{n}$, its Gauss-Seidel iterative scheme can be easily designed as
\begin{equation}
\left\{ {\begin{split}
	&\left( {{\gamma _3} + 4{\gamma _4}} \right)n_{1i,j}^{k + 1,l + 1} = {\gamma _4}\left( \text{U}\left(\bm n^{k+1,l}\right)-4n_{2i,j}^{k + 1,l} \right) - F_{i,j}^{k + 1,l + 1},n_{1i,j}^{k + 1,0} =n_{1i,j}^k\\
	&\left( {{\gamma _3} + 4{\gamma _4}} \right)n_{2i,j}^{k + 1,l + 1} = {\gamma _4}\left( \text{U}\left(\bm n^{k+1,l}\right)-4n_{1i,j}^{k + 1,l+1} \right) - F_{i,j}^{k + 1,l + 1},n_{2i,j}^{k + 1,0} =n_{2i,j}^k\\
	&\text{U}\left(\bm n^{k+1,l}\right)=\bm{n}_{i+1,j}^{k + 1,l}+\bm{n}_{i-1,j}^{k + 1,l}+\bm{n}_{i,j+1}^{k + 1,l}+\bm{n}_{i,j-1}^{k + 1,l}
	\end{split}} \right..
\end{equation}
Here, $\bm{n}$ can be solved with FFT as well.

For $\bm m$ in (3.23), its discretized analytical solution with projection formula is
\begin{equation}
\left\{ \begin{split}
&\bm {\tilde m_{i,j}}^{k + 1} = \bm n_{i,j}^{k + 1} + \frac{{\left( {\lambda _{1i,j}^k + {\gamma _1}} \right)\bm p_{i,j}^{k + 1} + \bm \lambda _{3i,j}^k}}{{{\gamma _3}}}\\
&\bm m_{i,j}^{k + 1} = \frac{{\bm {\tilde m_{i,j}}^{k + 1}}}{{\text{max}\left( {1,\left| {\bm {\tilde m_{i,j}}^{k + 1}} \right|} \right)}}
\end{split} \right..
\end{equation}

The $q$ obtained in (3.24) can also be drawn into a simple analytical solution
\begin{equation}
\left( {{\gamma _4} + 2b\left| {\bm p _{i,j}^{k + 1}} \right|{\delta _\varepsilon }\left( {\phi _{i,j}^{k + 1}} \right)} \right)q = {\gamma _4}\nabla  \cdot \bm n _{i,j}^{k + 1} - \lambda _{4i,j}^{k + 1}.
\end{equation}
After calculating (3.18) - (3.24), the Lagrange multipliers are updated as (3.10).

In each iteration, the following error tolerances should be checked to determine convergence, i. e.,
\begin{equation}
T_s^{k + 1} \le \text{Tol},\;\left( {s = 1,2,3,4} \right),\;{\Phi ^{k + 1}} \le \text{Tol},\;{\Sigma ^{k + 1}} \le \text{Tol},
\end{equation}
where $\text{Tol}=0.01$. $T_s^{k + 1},\;\Phi ^{k + 1},\;\Sigma ^{k + 1}$ are defined as
\begin{equation}
\begin{array}{lc}
\left\{
\begin{split}
{T_1^{k + 1}},\;{T_2^{k + 1}},\;{T_3^{k + 1}},\;{T_4^{k + 1}} 
\end{split}\right\}
= \left\{ {\begin{split}
	{\frac{{\| {\lambda _1^{k + 1} - \lambda _1^k} \|}_{L_1}}{{{{\| {\lambda _1^k} \|}_{{L_1}}}}}},\;{\frac{{{{\| {\bm \lambda _2^{k + 1} - \bm \lambda _2^k} \|}_{{L_1}}}}}{{{{\| {\bm \lambda _2^k} \|}_{{L_1}}}}}},\;{\frac{{{{\| {\bm \lambda _3^{k + 1} - \bm \lambda _3^k} \|}_{{L_1}}}}}{{{{\| {\bm \lambda _3^k} \|}_{{L_1}}}}}},\;{\frac{{{{\| {\lambda _4^{k + 1} - \lambda _4^k} \|}_{{L_1}}}}}{{{{\| {\lambda _4^k} \|}_{{L_1}}}}}}
	\end{split}} \right\},
\end{array}
\end{equation}
\begin{equation}
{\Phi ^{k + 1}} = \frac{{\| {\phi _{}^{k + 1} - \phi _{}^k} \|}_{{L_1}}}{{\| {\phi _{}^k} \|}_{{L_1}}},\;{\Sigma ^{k + 1}} = \frac{{\| {{E^{k + 1}} - {E^k}} \|}}{{\| {{E^k}} \|}}.
\end{equation}
The complete algorithm is summarized in Algorithm 1.

\hypertarget{alogrithm1}{}
\begin{table}[h!] 
	\centering
	\begin{tabular}{p{10cm}}
		\hline
		\rowcolor{Gray}
		\textbf{Algorithm 1}: ADMM \\
		\hline
		1: Initialization: Set ${\rm{ }}{\alpha _1},\;{\alpha _2},\;\mu ,\;a,\;b$.\\
		2: \textbf{while} \textit{any stopping criterion is not satisfied } \textbf{do}\\
		\qquad Calculate ${\rm{ c}}_1^{k + 1},\;{\rm{c}}_2^{k + 1}{\rm{ }}$ from (3.18) and (3.19)\\
		\qquad Calculate ${\phi ^{k + 1}}$ from (3.20) \\
		\qquad Calculate ${\bm{p} ^{k + 1}}$ from (3.21) \\
		\qquad Calculate ${\bm{n} ^{k + 1}}$ from (3.22) \\
		\qquad Calculate ${\bm{m} ^{k + 1}}$ from (3.23) \\
		\qquad Calculate ${q ^{k + 1}}$ from (3.24) \\
		\qquad Calculate $\lambda _1^{k + 1},\bm \lambda _2^{k + 1},\bm \lambda _3^{k + 1},\lambda _4^{k + 1}$ from ${\rm{ }}(3.10)$  \\
		3: \textbf{end while}\\
		\hline
	\end{tabular}
\end{table} 

\section{Numerical Experiments}
In this section, we devise four sets of numerical experiments with distinct goals in mind. The first set compares the performance of the CV, the CVL, the CVE, and the CVEL model in contour inpainting or interpolation, in cases where smaller and larger regions are missing from the original images. The second set of experiments examines the dependence of the segmentation result on the number of landmark points. The third one demonstrates how landmark points improve the segmentation efficiency of the CVEL model, and the fourth one is an  application in medical image segmentation. 
\subsection{Comparisons with previous models}
Since the CVE, CVL, and CVEL model are different extensions of the CV model for the purpose of missing contour recovery, we design some experiments to compare their performance. 
First, we segment an image of the letters 'UCLA' with small damaged regions \cite{zhu2013image}. We give the original broken image in Fig.1a, and the initialization of the zero level set for all of the models in Fig.1b. Segmentation results obtained via the CV, CVL, CVE, and CVEL are shown in Fig.1c, Fig.1d, Fig.1e, and Fig.1f respectively. The parameters used in the CVEL model are $\gamma_1=1,\gamma_2=3,\gamma_3=5,\gamma_4=10,\alpha_1=0.5,\alpha_2=0.5$. One landmark was placed in the middle of each piece of missing contour. Results show that the CVL, CVE, and CVEL can all recover small sections of the missing contours, though the CVEL produces smoother curves due to the elastica regularizer.

\begin{figure}[h!] 
	\centering  
	\subfigure[]{\includegraphics[width=0.3\textwidth]{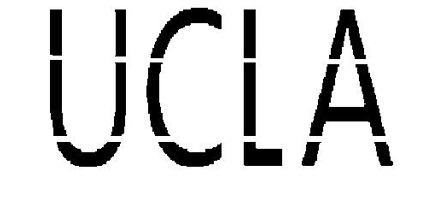}}
	\subfigure[]{\includegraphics[width=0.3\textwidth]{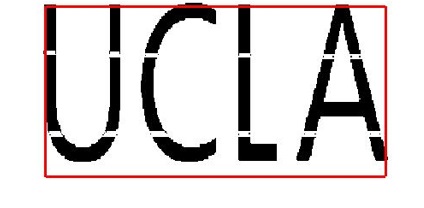}}
	\subfigure[]{\includegraphics[width=0.3\textwidth]{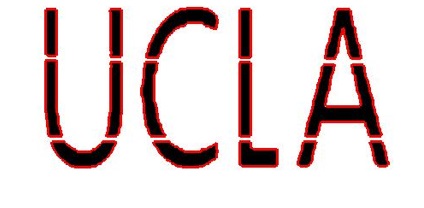}}
	\subfigure[]{\includegraphics[width=0.3\textwidth]{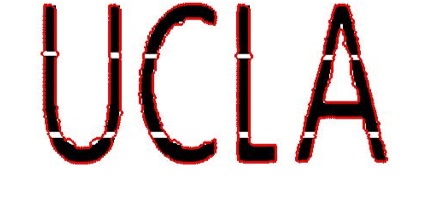}}
	\subfigure[]{\includegraphics[width=0.3\textwidth]{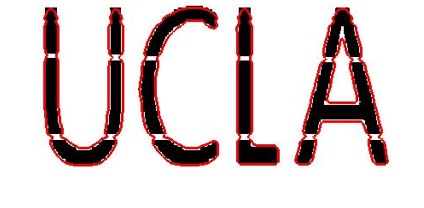}}
	\subfigure[]{\includegraphics[width=0.3\textwidth]{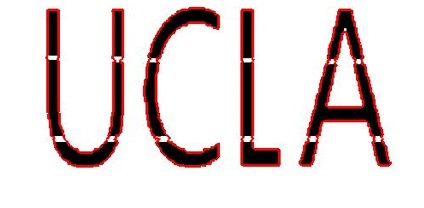}}
	\vspace{-10pt}
	\caption{Results of four different methods of repair broken letters 'UCLA'}
	\label{fig:residuals}
\end{figure}
Next, to compare the performance of the CVL and CVEL in recovering larger missing contours, we conducted the second experiment shown in Fig.2. Fig.2a shows a triangle with a missing corner, Fig.2b shows the initial contour and landmark points, Fig.2c and Fig.2d are segmentation results via the CVL and CVEL model respectively, and the parameters are $\gamma_1=1,\gamma_2=3,\gamma_3=5,\gamma_4=10,\alpha_1=1.1,\alpha_2=0.9$ for both models. Although they obtained similar results, the CVL needed 26 landmark points whereas the CVEL needed only 20. 

\begin{figure}[h!] 
	\centering  
	\subfigure[]{\includegraphics[width=0.24\textwidth]{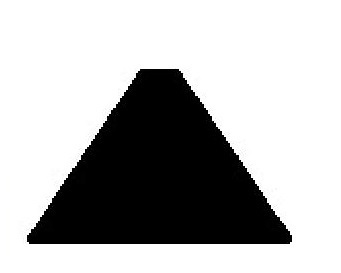}}
	\subfigure[]{\includegraphics[width=0.24\textwidth]{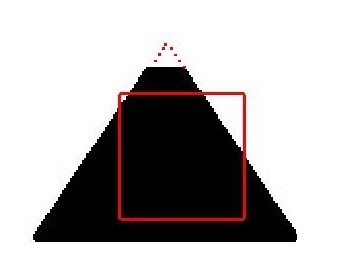}}
	\subfigure[]{\includegraphics[width=0.24\textwidth]{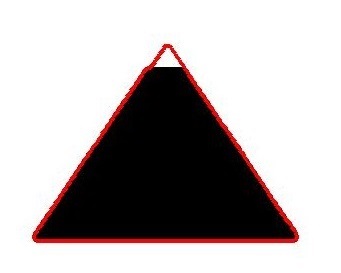}}
	\subfigure[]{\includegraphics[width=0.24\textwidth]{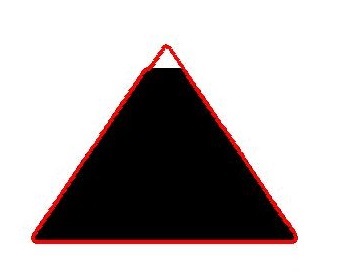}}\\
	\vspace{-10pt}
	\caption{Broken triangle repair experiment}
	\label{fig:residuals}
\end{figure}
Furthermore, we set up an experiment to compare the performance of the CVL and CVEL, as shown in Fig.3. Fig.3a shows the original broken image, Fig.3b presents the initial zero level set and landmark points, and Fig.3c and Fig.3d give the segmented results via CVL and CVEL model respectively.
The parameters of CVEL model are $\gamma_1=7,\gamma_2=20,\gamma_3=5,\gamma_4=2,\alpha_1=1.1,\alpha_2=0.9$. The conclusion is the same as \cite{zhu2013image}, i. e., it is hard to inpaint the external missing curve via the CVL, but the CVEL works well. This is mainly because the curvature term has good curve repair properties. 
\begin{figure}[h!] 
	\centering  
	\subfigure[]{\includegraphics[width=0.24\textwidth]{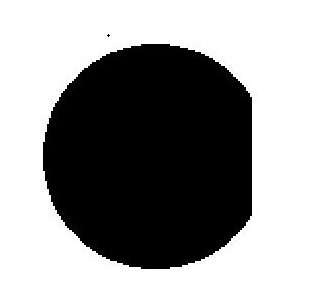}}
	\subfigure[]{\includegraphics[width=0.24\textwidth]{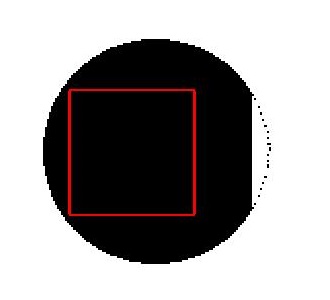}}
	\subfigure[]{\includegraphics[width=0.24\textwidth]{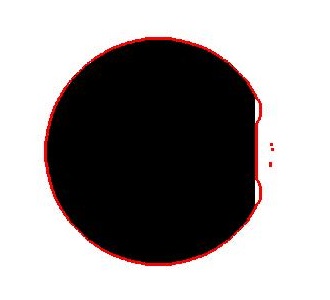}}
	\subfigure[]{\includegraphics[width=0.24\textwidth]{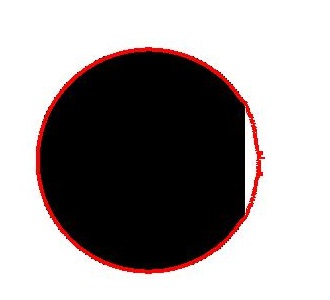}}\\
	\vspace{-10pt}
	\caption{Broken circle repair experiment.}
	\label{fig:residuals}
\end{figure}

\subsection{The dependence on tuning parameters and landmark points}
In this part of experiments, we study the effect of the number of landmark points and their positions on the segmentation result.
Setting different amounts of landmark points lead to different results. The more landmarks we set within a certain limit, the more accurate the result tends to be. However, increasing the number of landmarks beyond this limit will not increase segmentation accuracy, as shown in Fig.4.a to Fig.4d. In Fig.4e to Fig.4h, the number of landmark points is 2, 10, 18, 24, respectively. As we can see, the corresponding results became increasingly better with the more and more additional landmarks. However, setting over 24 landmarks did not improve the result further.

\begin{figure}[h!] 
	\centering  
	\subfigure[]{\includegraphics[width=0.18\textwidth]{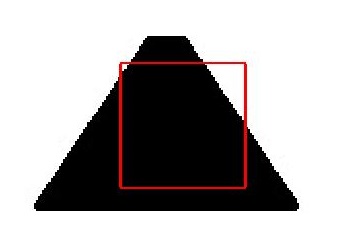}}
	\subfigure[]{\includegraphics[width=0.2\textwidth]{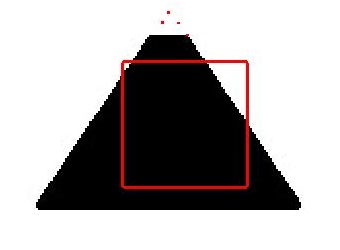}}
	\subfigure[]{\includegraphics[width=0.2\textwidth]{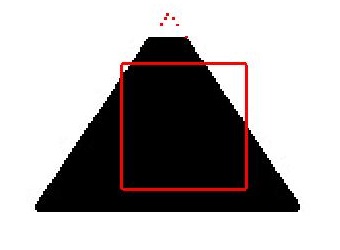}}
	\subfigure[]{\includegraphics[width=0.2\textwidth]{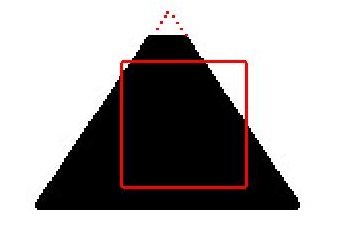}}\\
	\subfigure[]{\includegraphics[width=0.2\textwidth]{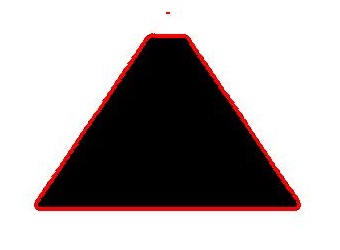}}
	\subfigure[]{\includegraphics[width=0.2\textwidth]{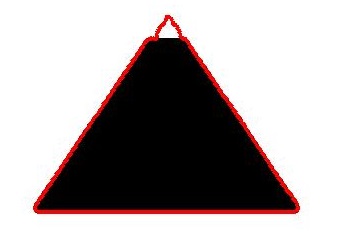}}
	\subfigure[]{\includegraphics[width=0.2\textwidth]{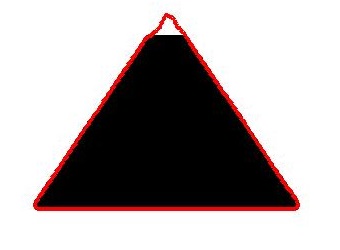}}
	\subfigure[]{\includegraphics[width=0.2\textwidth]{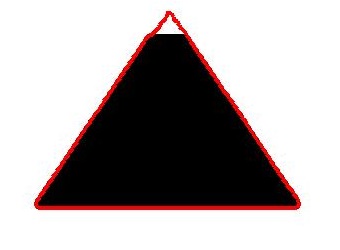}}\\
	\vspace{-10pt}
	\caption{Different number of landmark points affect the results of the experiment}
	\label{fig:residuals}
\end{figure}

The placement of landmark points is also essential, especially when the total number of landmarks is small. Using Fig. 4d as an example of a well-segmented image, we proceed to take away landmarks from different locations. In Fig.5a, Fig.5b and Fig.5c, we remove two landmarks from the bottom, top, and middle of the broken tip of the triangle, respectively. As a result, the recovered contour in Fig.5d has distortions around the base section, the sharpness of the tip is not well maintained in Fig.5e, and the result in Fig.5f does not change significantly. Therefore, we observe that it is more effective to place landmarks at the vertices or corners of an object. The better the landmark positions, the fewer landmarks we need.

\begin{figure}[h!] 
	\centering  
	\subfigure[]{\includegraphics[width=0.25\textwidth]{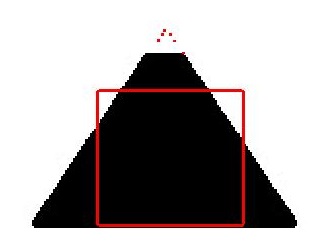}}
	\subfigure[]{\includegraphics[width=0.25\textwidth]{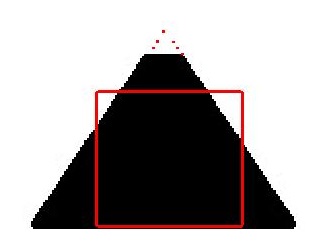}}
	\subfigure[]{\includegraphics[width=0.25\textwidth]{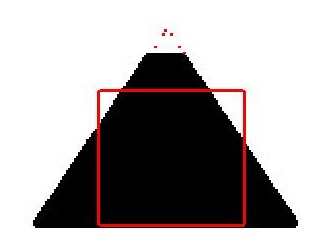}}\\
	\subfigure[]{\includegraphics[width=0.25\textwidth]{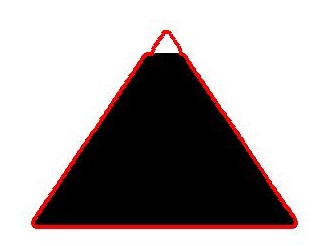}}
	\subfigure[]{\includegraphics[width=0.25\textwidth]{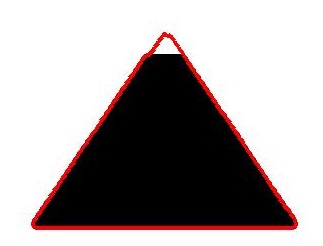}}
	\subfigure[]{\includegraphics[width=0.25\textwidth]{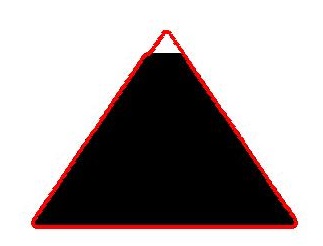}}\\
	\vspace{-10pt}
	\caption{Experiment on the effect of landmark points in different locations on the results}
	\label{fig:residuals}
\end{figure}

\subsection{Efficiency}
In this section, we examine how the CVEL performs in terms of efficiency, specifically the convergence time.
We first consider whether the CVEL speeds up segmentation by constructing the experiment in Fig 6, where we mark the entire contour of the palm in the CVEL and compare performance with the CVE. Fig.6a is the original image, and Fig.6b is the initial zero level set for both models. We obtain the segmentation results shown in Fig.6c and Fig.6d via CVE by five steps and CVEL by two steps, respectively. This shows that using landmarks in the CVEL model can increase efficiency.

\begin{figure}[h!] 
	\centering  
	\subfigure[]{\includegraphics[width=0.2\textwidth]{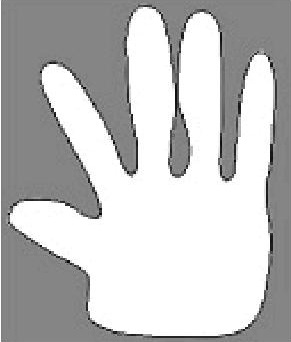}}
	\subfigure[]{\includegraphics[width=0.2\textwidth]{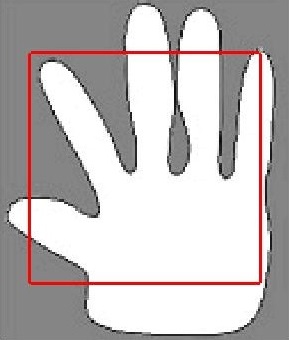}}
	\subfigure[]{\includegraphics[width=0.2\textwidth]{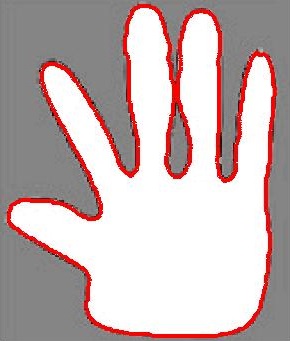}}
	\subfigure[]{\includegraphics[width=0.2\textwidth]{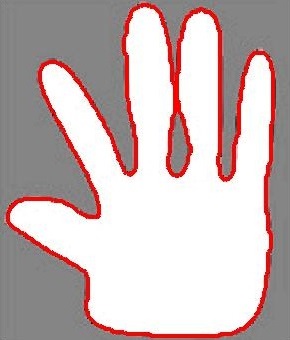}}\\
	\vspace{-10pt}
	\caption{Hand image segmentation efficiency test}
	\label{fig:residuals}
\end{figure}
\begin{figure}[h!] 
	\centering  
	\subfigure[]{\includegraphics[width=0.3\textwidth]{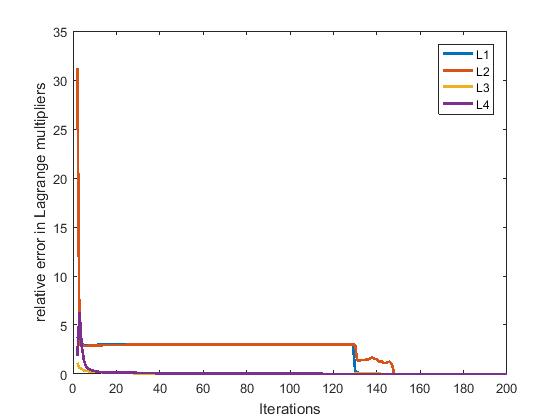}}
	\subfigure[]{\includegraphics[width=0.3\textwidth]{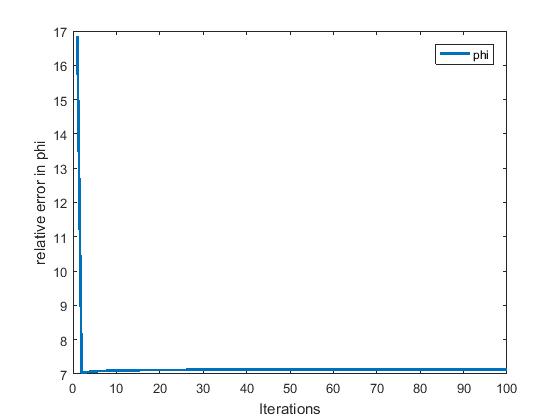}}
	\subfigure[]{\includegraphics[width=0.3\textwidth]{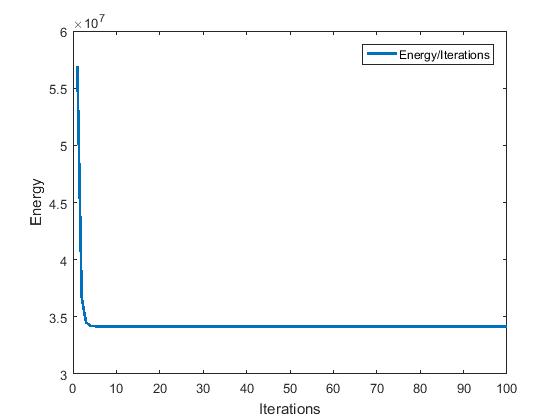}}\\
	\subfigure[]{\includegraphics[width=0.3\textwidth]{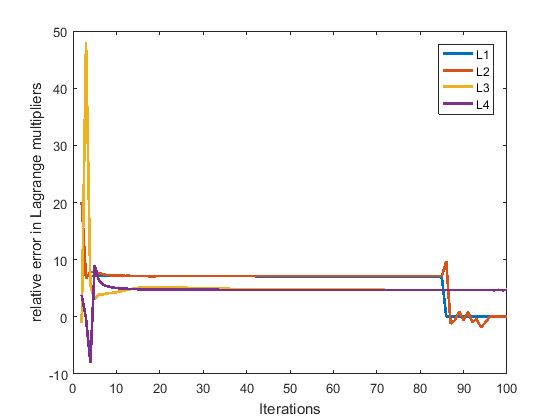}}
	\subfigure[]{\includegraphics[width=0.3\textwidth]{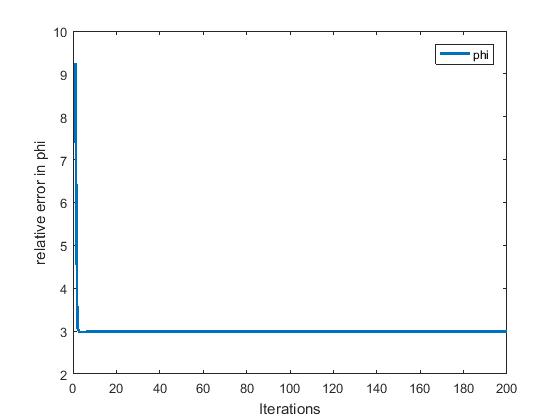}}
	\subfigure[]{\includegraphics[width=0.3\textwidth]{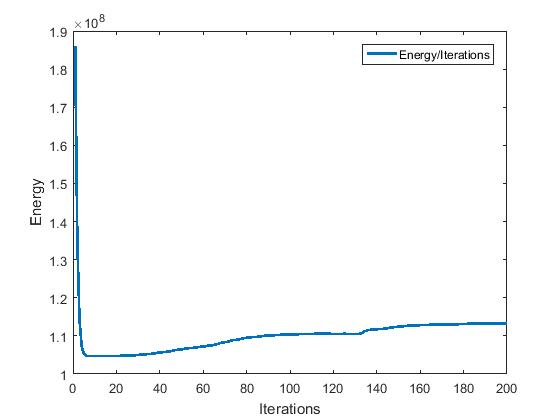}}\\
	\vspace{-10pt}
	\caption{The plots of relative errors in Lagrange multipliers, relative error in phi, and energy for the two examples “hand” and “UCLA”. The first row lists the plots for “hand” and the second one for “UCLA”.}
	\label{fig:residuals}
\end{figure}

We then check the convergence time of the Lagrangian multiplier, the level set function, and the total energy (4.16-4.17) using the CVEL model, where Fig.7a, Fig.7b, and Fig.7c map their respective values in the experiment in Fig.6, and Fig.7d, Fig.7e, and Fig.7f map the same values in the experiment in Fig.1.  On the one hand, we see that convergence is reached quickly in both experiments. On the other hand, we observe that the total energy increased towards the end in Fig.7f. This phenomenon is due to the model setting the landmark points onto an illusory contour. In the initial stage, the CV and elastica terms play a major role in moving the curve towards the broken boundary. As the curve approaches the landmarks, it moves away from the natural boundary as indicated by the separation of regions according to image information. This process inevitably raises the total energy.

\subsection{Applications in real life}
In this sub-section, we present some applications of the CVEL in segmenting CT images which often encounter difficulties due to the presence of fine details. 

\begin{figure}[h!] 
	\centering  
	\subfigure[]{\includegraphics[width=0.18\textwidth]{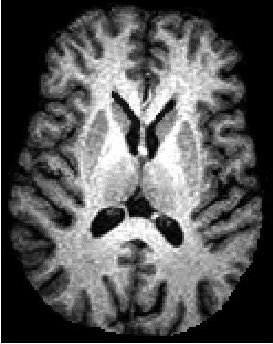}}
	\subfigure[]{\includegraphics[width=0.18\textwidth]{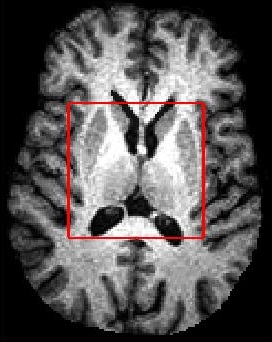}}
	\subfigure[]{\includegraphics[width=0.18\textwidth]{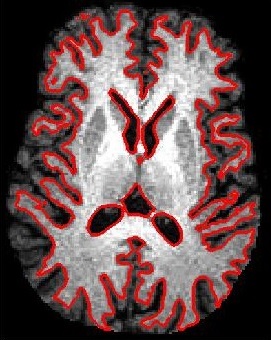}}
	\subfigure[]{\includegraphics[width=0.18\textwidth]{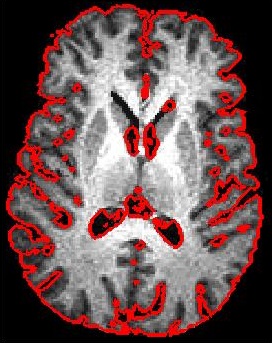}}\\
	\subfigure[]{\includegraphics[width=0.18\textwidth]{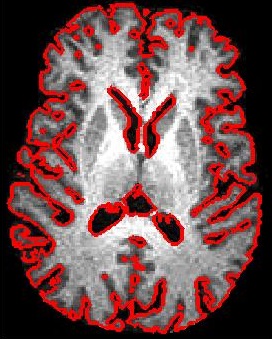}}
	\subfigure[]{\includegraphics[width=0.18\textwidth]{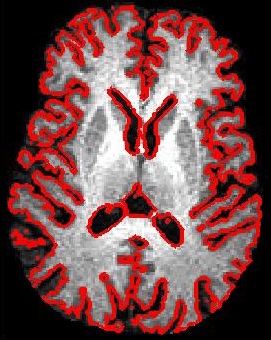}}
	\subfigure[]{\includegraphics[width=0.18\textwidth]{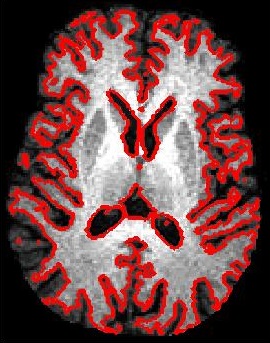}}
	\subfigure[]{\includegraphics[width=0.18\textwidth]{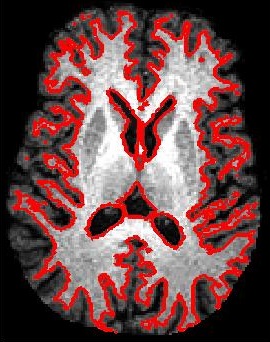}}\\
	\vspace{-10pt}
	\caption{Classical brain CT image segmentation experiments}
	\label{fig:residuals}
\end{figure}

Fig.8a gives an original CT image of brain, Fig8.b presents the initial level set function for segmentation, Fig8.c shows the result via CV model which fails to separate the brain tissue completely. On the contrary, the CVEL model with proper landmark points and using the same level set initialization can produce good result as presented in Fig8.e. Since the missing parts in Fig8.c are areas with low image intensities, We attempt to improve quality through adding landmarks in these areas. The results in Fig8.d-Fig.8h obtained via marking points over intensity of 20,40,60,80,100, but Fig8.d presents over segmentation result due to redundant information, and Fig8.f-Fig.8h present deficient segmentation results due to less mark points. 

\begin{figure}[h!] 
	\centering  
	\subfigure[]{\includegraphics[width=0.2\textwidth]{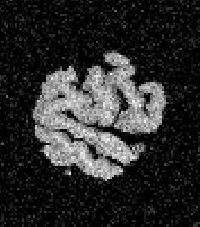}}
	\subfigure[]{\includegraphics[width=0.2\textwidth]{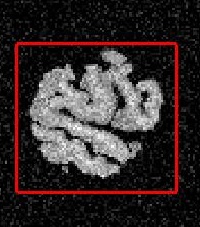}}
	\subfigure[]{\includegraphics[width=0.2\textwidth]{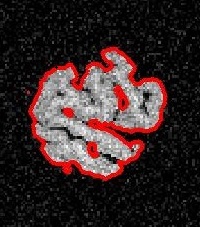}}
	\subfigure[]{\includegraphics[width=0.2\textwidth]{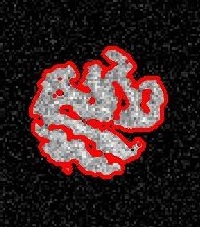}}\\
	\vspace{-10pt}
	\caption{Classical brain CT image with noises segmentation experiments}
	\label{fig:residuals}
\end{figure}
Fig.9 gives another example of CT image segmentation with noises. The main difficulty in this experiment is to segment the adjacent areas in the image. For the original image Fig.9a, we initialize the level set function as Fig.9b. The Fig.9c and Fig.9d show the results obtained via CV model and CVEL model respectively. While using CVEL, we add landmark points over 100 of intensity for a satisfactory result.
\section{Concluding remarks}
In this paper, we present a Chan-Vese model with elastica and landmarks (CVEL) under the variational level set framework. The new model combines the classical Chan-Vese model (CV), the Chan-Vese model with landmarks (CVL), and the Chan-Vese model with elastica (CVE). We then design an ADMM algorithm for the solution. A variety of numerical experiments show that the CVEL performs better than the CVE in segmentation accuracy, and can recover larger broken boundaries than the CVL.For future work, we can integrate automatic landmarks detection methods as well as other techniques involving priors and ultimately aim to achieve automatic segmentation.\\ 
\section{Acknowledgement}
The authors thank the editor and anonymous reviewers for their helpful
comments and valuable suggestions.
\bibliographystyle{unsrt}
\bibliography{Tensor}

\end{document}